\documentclass[11pt,a4paper]{article}
\usepackage[hyperref]{naaclhlt2019}
\usepackage{times}
\usepackage{latexsym}
\usepackage{graphicx}  
\usepackage{amssymb}
\usepackage{amsmath}
\usepackage{multirow}
\usepackage{color}
\usepackage{hyperref}

\usepackage{url}

\aclfinalcopy 

\title{Attentional Multi-Reading Sarcasm Detection}
\author{
Reza Ghaeini, Xiaoli Z. Fern, Prasad Tadepalli\\
School of Electrical Engineering and Computer Science, Oregon State University\\
	1148 Kelley Engineering Center, Corvallis, OR 97331-5501, USA\\
	{\tt \{ghaeinim, xfern, tadepall\}@eecs.oregonstate.edu}\\
 }

\date{}

\begin{document}
\maketitle

\begin{abstract}
Recognizing sarcasm often requires a deep understanding of multiple sources of information, including the utterance, the conversational context, and real world facts. Most of the current sarcasm detection systems consider only the utterance in isolation.  There are some limited attempts toward taking into account the conversational context. In this paper, we propose an interpretable end-to-end model that combines information from both the utterance and the conversational context to detect sarcasm, and demonstrate its effectiveness through empirical evaluations. We also study the behavior of the proposed model to provide explanations for the model's decisions. Importantly, our model is capable of determining the impact of utterance and conversational context on the model's decisions. Finally, we provide an ablation study to illustrate the impact of different components of the proposed model.
\end{abstract}

\section{Introduction}
Recently, dialogue systems have received a lot of attention from researchers. Unfortunately, existing approaches often fail to detect sarcastic user comments in order to provide proper responses.

Sarcasm detection is an important and challenging task for natural language understanding. The goal of sarcasm detection is to determine whether a sentence is sarcastic or non-sarcastic. Sarcasm is a type of phenomenon with specific perlocutionary effects on the hearer \cite{haver}, such as to break their pattern of expectation. Consequently, correct understanding of sarcasm often requires a deep understanding of multiple sources of information, including the utterance, the conversational context, and, frequently some real world facts. Table~\ref{tab:sample} shows three different sarcastic samples from the SARC dataset \cite{sarc}, each of which requires a different source of information for disambiguation.

\begin{table}[t]
	\small
	\begin{center}
		\begin{tabular}{c|c|l}
			\hline 
            Type & \multicolumn{2}{c}{Sample} \\ \hline
			\multirow{4}{*}{U.S.$^a$} & \multirow{3}{*}{C$^d$} & just don't. if you are telling anyone else \\ 
			& & what they can and can't put on their \\ 
			& & bodies, just don't \\ \cline{2-3}
            & \multirow{2}{*}{R$^e$} & \textcolor{blue}{we're on Reddit}, don't you know \textcolor{blue}{we}  \\ 
            & & \textcolor{blue}{control everything people do}? \\ \hline
            
			\multirow{4}{*}{C.D.$^b$} & \multirow{3}{*}{C} & who else thinks that \textcolor{blue}{javascript alert} is \\
            & & an \textcolor{blue}{annoying, lazy, and ugly way} to \\
            & & notify  me of something on your site. \\ \cline{2-3}
			& R & it's a \textcolor{blue}{useful debugging tool} \\ \hline
            
			\multirow{9}{*}{E.K.D.$^c$} & \multirow{6}{*}{C} & till that some cattle ranchers in south \\ 
            & & dakota lost between 20\% - 50\% of their \\ 
            & & livestock in winter storm atlas, and may  \\ 
            & & not be eligible for insurance due to the   \\ 
            & & expiration of the farm bill and  \\ 
            & & federal government shutdown.\\ \cline{2-3}
            & \multirow{3}{*}{R} & this is clearly \textcolor{blue}{barrack hussein obama's} \\ 
            & & fault, since he refuses to modify the \\ 
            & & \textcolor{blue}{aca} and \textcolor{blue}{obamacare}. \\ \hline
			\multicolumn{3}{l}{$^a$\textbf{U.S.}, Utterance Sufficient.}\\
            \multicolumn{3}{l}{$^b$\textbf{C.D.}, Conversation Dependent.}\\
			\multicolumn{3}{l}{$^c$\textbf{E.K.D.}, External Knowledge Dependent.}\\ 
            \multicolumn{3}{l}{$^d$\textbf{C}, Comment.} \\
            \multicolumn{3}{l}{$^e$\textbf{R}, Response.}\\ \hline
            


		\end{tabular}
	\end{center}
	\caption{\label{tab:sample} Different types of sarcastic examples from the SARC dataset. Each data sample contains a comment and response. Important and influential tokens are shown in blue.}
\end{table}

Existing approaches for sarcasm detection primarily focus on lexical, pragmatic cues (e.g. interjections, punctuations, sentimental shift etc.) found in utterance \cite{lexical,sarc_survey}. In contrast, the natural language understanding aspect of sarcasm detection could be more robust, interesting and challenging. Moreover, most sarcasm detection systems have considered utterances in isolation \cite{davidov,gonzalez,liebrecht,rilof13,maynard,joshi15,ghosh15,joshi16,ghosh16,poria_sarc,amir_2016,cascade}. However, even humans have difficulty in recognizing sarcastic intent when considering an utterance in isolation \cite{wallace_data}. There are some limited attempts toward taking the conversational context into account \cite{conv_sarc} by using a variety of LSTMs \cite{lstm} to encode both context and reply sentences. Still such approaches only focuses on the conversation dependent samples.

In this work, we propose an end-to-end model that combines information from both the utterance and the conversational context to detect sarcasm. Considering the utterance beside the conversational context enables the model to (1) properly handle utterance-sufficient samples, (2) automatically extract lexical and grammatical features from the utterance. First, We demonstrate the effectiveness of our model through empirical evaluations on the SARC dataset \cite{sarc}, the largest available dataset for sarcasm detection. Next, we illustrate the impact of different aspects of the proposed model through an ablation study. Finally, we present an extensive data analysis to (1) provide explanations regarding our model's decisions and behavior by visualizing attention and attention saliency\cite{sali_vis}; (2) study the impact and effect of utterance and the conversational context on our model's final prediction.
In summary, our contributions are as follows: 
\begin{itemize}
\item Proposing a novel end-to-end and interpretable deep learning model that combines information from both the utterance and conversational context in parallel.
\item Illustrating the impact of the proposed model's component through an extensive ablation study.
\item Explaining the model's behavior and predictions by visualization of the attention and attention saliency.  
\item Examining the impact of utterance and conversational context on the model's final predictions.
\end{itemize} 

\section{Related Work}

Automatic sarcasm detection is a relatively recent field of research. Early studies use small datasets and leverage lexical and syntactic features for sarcasm detection \cite{sarc_survey}. Here we classify the previous works into three categories, isolate-utterance based, contextual-feature based, and conversation based sarcasm detection models.
\begin{itemize}

\item \textbf{Isolate-utterance based:} Most existing sarcasm detection systems consider the utterances in isolation~\cite{davidov,gonzalez,liebrecht,rilof13,maynard,joshi15,ghosh15,joshi16,ghosh16}. Methods in this category commonly rely on hand-designed features, syntactic patterns, and lexical cues. 

\item \textbf{Contextual-feature based:} \citet{wallace_data} illustrates the necessity of using contextual information in sarcasm detection by showing how traditional classifiers fail in instances where humans also require additional context. Consequently, researchers recently started to exploit contextual information for sarcasm detection. In particular, contextual information about authors, topics or conversational context have been considered \cite{khattri15,bamman15,wallace_data2,rajadesingan15,poria_sarc,zhang16,amir_2016,cascade}. Such techniques rely on either feature engineering or embedding-based representation via deep learning.

These approaches benefit from contextual information in a pipelined and feature based manner. 
We should note that user profiling has been shown to have noticeable impact on sarcasm detection \cite{cascade}. However, user profiling is not always possible. In this work, we are primarily interested in the language side of the sarcasm detection and aim to provide an end-to-end user/author independent system that could be used in a variety of applications, especially dialogue systems and chat boxes.

\item \textbf{Conversation-based:} 
The last category of methods aims to detect sarcasm based on the understanding of the conversation (other than simply extracting features from the context). To the best of our knowledge, there is just one conversation dependent sarcasm detection system \cite{conv_sarc}, which focuses on modeling conversational context using a variety of LSTMs to help sarcasm detection. They effectively demonstrated the importance and impact of considering conversational context for sarcasm detection.

\end{itemize}

\noindent Among all previous works, \citet{conv_sarc} and our system share similar intuition and motivation. However, we utilize a different deep learning architecture to address sarcasm detection. Furthermore, we consider the utterance in both isolation and conversation dependent settings. Such a strategy allows the model to (1) extract lexical and grammatical features from the utterance, and (2) selectively attend to the proper source of information. Finally, we evaluate our system with a much larger and broader dataset that could lead to more robust and unbiased evaluation.

\section{Model}
	The inputs to our model are $u=[u_1, \cdots, u_n]$ and $v=[v_1, \cdots, v_m]$, which are the given comment (length $n$) and response (length $m$) respectively. Here $u_i, v_j \in \mathbb{R}^r$ are $r$-dimensional word embedding vectors. The goal is to predict a label $y$ that indicates whether the response $v$ is sarcastic or non-sarcastic.

\begin{figure}[t]
	\centering
	\includegraphics[width=.44\textwidth]{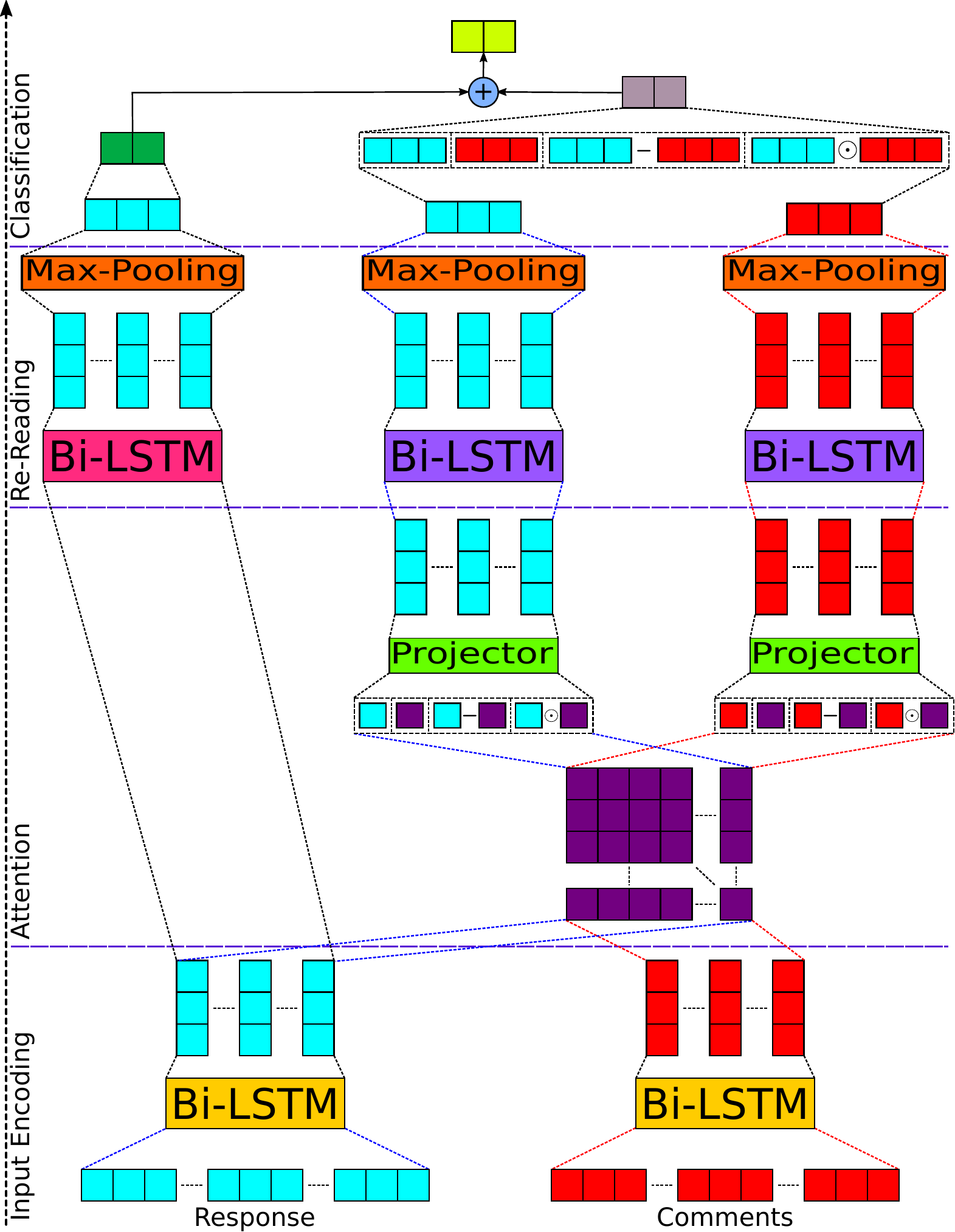}
	\caption{A high-level view of our model (AMR). The data (comment $u$ and response $v$, depicted with red and cyan/blue tensors respectively) flows from bottom to top. Relevant tensors are shown with the same color and elements with the same colors share parameters. The left part shows the utterance-only part and the right part represents the conversation-dependent part of AMR. \label{fig:model}}
\end{figure}

	Our proposed model (\textbf{A}ttentional \textbf{M}ulti-\textbf{R}eading system; AMR) consists of an utterance-only (left side) part and a conversation-dependent (right side) part, formulated with the following major components: input encoding, attention, re-reading, and classification. Figure~\ref{fig:model} demonstrates a high-level view of our proposed AMR framework. 
    
\subsection{Input Encoding}
	\label{sec:enc}
	
	RNNs provide a natural solution for modeling variable length sequences and have shown to be successful in various NLP tasks \cite{dgr,drbilstm,nmt,event}. Consequently, we utilize a bidirectional LSTM (BiLSTM) \cite{lstm} for encoding the given comment and response. Here we simply read and encode the comment and response using a BiLSTM. Equations \ref{eq:enc:u} and \ref{eq:enc:v} formally represent this component.

\begin{equation}
	\bar{u} = \textit{BiLSTM}(u)
	\label{eq:enc:u}
\end{equation}
\vspace{-.1in}
\begin{equation}
	\bar{v} = \textit{BiLSTM}(v)
	\label{eq:enc:v}
\end{equation}
	
	\noindent where $\bar{u} \in \mathbb{R}^{n \times 2d}$ and $\bar{v} \in \mathbb{R}^{m \times 2d}$ are the BiLSTM reading sequences of $u$ and $v$ respectively.

\subsection{Attention}
    Here we employ a soft alignment method to associate the relevant sub-components between the given comment and response. 
The unnormalized attention weights are computed as the similarity of the hidden states of the comment and response as shown in Equation~\ref{eq:energy} (energy function).
    
\begin{equation}
	e_{ij} = \bar{u}_i \bar{v}_j^T,  \quad  i \in [1,n], j \in [1,m]
	\label{eq:energy}
\end{equation}
	
	\noindent where $\bar{u}_i$ and $\bar{v}_j$ are the hidden representations of $u$ and $v$ respectively which are computed earlier in Equations \ref{eq:enc:u} and \ref{eq:enc:v} respectively. Next, for each word in either comment or response, the relevant semantics in the other sentence is extracted and composed according to $e_{ij}$ as shown in Equations \ref{eq:att:c} and \ref{eq:att:r}.
	
\begin{equation}
	\tilde{u}_i = \sum_{j=1}^{m} \frac{\exp(e_{ij})}{\sum_{k=1}^{m} \exp(e_{ik})} \bar{v}_j, \quad i \in [1,n]
	\label{eq:att:c}
\end{equation}

\begin{equation}
	\tilde{v}_j = \sum_{i=1}^{n} \frac{\exp(e_{ij})}{\sum_{k=1}^{n} \exp(e_{kj})} \bar{u}_i, \quad j \in [1,m]
	\label{eq:att:r}
\end{equation}
	
\noindent where $\tilde{u}_i$ represents the extracted relevant information of $\bar{v}$ by attending to $\bar{u}_i$ while $\tilde{v}_j$ represents the extracted relevant information of $\bar{u}$ by attending to $\bar{v}_j$. 
	
\subsubsection{Attention Augmentation and Projection}
To utilize the collected attentional information $\tilde{u}_j$ and $\tilde{v}_j$, a trivial next step would be to concatenate them with $\bar{u}_i$ and $\bar{v}_j$ respectively. More over, it is often interesting to compare and contrast the information from the comment and the response in order to detect sarcasm. Hence, we calculate the element-wise difference and element-wise and include these vectors for further consideration.    
We concatenate all the vectors and represent the comment and response as $[\bar{u}_i, \tilde{u}_i, \bar{u}_i -\tilde{u}_i, \bar{u}_i \odot \tilde{u}_i]$ and 	$[\bar{v}_j, \tilde{v}_j, \bar{v}_j -\tilde{v}_j, \bar{v}_j \odot \tilde{v}_j]$ with $i=1,...,n$ and $j=1,...,m$ respectively. 
Finally, a feed-forward neural layer with the ReLU activation function projects the concatenated vectors from the $8d$-dimensional vector space into a $d$-dimensional vector space (Equations \ref{eq:prj:c} and \ref{eq:prj:r}). This projection layer serves the dual purpose of both helping the model to capture deeper dependencies between the comment and response and  lowering the complexity of vector representations.
\begin{equation}
	p_i =  \textit{ReLU}(W_c ([\bar{u}_i, \tilde{u}_i, \bar{u}_i -\tilde{u}_i, \bar{u}_i \odot \tilde{u}_i]) + b_c)
	\label{eq:prj:c}
\end{equation}
\vspace{-.2in}
\begin{equation}
    q_j =  \textit{ReLU}(W_c ([\bar{v}_j, \tilde{v}_j, \bar{v}_j -\tilde{v}_j, \bar{v}_j \odot \tilde{v}_j]) + b_c)
	\label{eq:prj:r}
\end{equation}
	
	\noindent Here $\odot$ stands for element-wise product while $W_c \in \mathbb{R}^{8d\times d}$ and $b_c \in \mathbb{R}^{d}$ are the trainable weights and biases of the projector layers respectively.
	
\subsection{Re-Reading}
	During this phase, two BiLSTMs are used. First, we use a shared BiLSTM ($\textit{BiLSTM}_c$) to aggregate the sequences of computed matching vectors, $p$ and $q$ from the \emph{Attention} stage. This aggregation is performed in a sequential manner to ensure that sequential information in the latent variables is retained. 
Second, We use another BiLSTM to re-read and re-encode the previous encoding of the response from the \emph{Input Encoding} section ($\bar{v}$). Such a re-reading process is helpful toward achieving a deeper and more meaningful representation for the response when considered in isolation. The Re-Reading procedure is done through Equations \ref{eq:re_enc:c}, \ref{eq:re_enc:r}, and \ref{eq:re_enc:self}.
	
\begin{equation}
	\bar{p} = \textit{BiLSTM}_c(p)
	\label{eq:re_enc:c}
\end{equation}
\vspace{-.1in}
\begin{equation}
	\bar{q} = \textit{BiLSTM}_c(q)
	\label{eq:re_enc:r}
\end{equation}
\vspace{-.1in}    
\begin{equation}
	\bar{x} = \textit{BiLSTM}_u(\bar{v}) 
	\label{eq:re_enc:self}
\end{equation}

	Finally, we convert $\bar{p} \in \mathbb{R}^{n\times 2d}$, $\bar{q} \in \mathbb{R}^{m \times 2d}$ and $\bar{x} \in \mathbb{R}^{m \times 2d}$ to fixed-length vectors using a max pooling layer (Equations \ref{eq:fixlpool:c}, \ref{eq:fixlpool:r}, and \ref{eq:fixlpool:self}).
	
\begin{equation}
	\tilde{p} = \textit{MaxPooling}(\bar{p}) 
	\label{eq:fixlpool:c}
\end{equation}
\vspace{-.1in}
\begin{equation}
	\tilde{q} = \textit{MaxPooling}(\bar{q})
	\label{eq:fixlpool:r}
\end{equation}
\vspace{-.1in}
\begin{equation}
	\tilde{x} = \textit{MaxPooling}(\bar{x})
	\label{eq:fixlpool:self}
\end{equation}

\noindent where $\tilde{p} \in \mathbb{R}^{2d}$, $\tilde{q} \in \mathbb{R}^{2d}$ are the final and fixed representations of the comment and the response produced via conversation-dependent reading (the right part of the model), and $\tilde{x} \in \mathbb{R}^{2d}$ is a separate representation of the response produced by the utterance-only reading (the left portion of the model).
	
\subsection{Classification}
To make final prediction, we consider both the utterance-only representation as well as the conversation dependent representations. 
Equation~\ref{eq:out:self} represents a feed-forward layer that computes the utterance-only prediction from $\tilde{x}$. For the conversation-dependent part, we enrich the extracted information from the comment and response by incorporating the difference and element-wise product of $\tilde{p}$ and 
$\tilde{q}$ respectively. Equation~\ref{eq:out:conv} formally describes the prediction procedure for the conversation-dependent part.

\begin{equation}
	o_u =  U_u \tilde{x} + a_u
	\label{eq:out:self}
\end{equation}
\vspace{-.1in}
\begin{equation}
	o_c =  U_c ([\tilde{p}, \tilde{q}, \tilde{p} -\tilde{q}, \tilde{p} \odot \tilde{q}]) + a_c
	\label{eq:out:conv}
\end{equation}

\noindent where $U_u \in \mathbb{R}^{2d\times 2}$, $U_c \in \mathbb{R}^{8d\times 2}$, $a_u \in \mathbb{R}^{2}$ and $a_c \in \mathbb{R}^{2}$ are the trainable weights and biases of the prediction layers respectively. 
Finally, we combine both predictions (i.e.~$o_u$ and $o_c$) using a trainable weight $\alpha$ (Equation~\ref{eq:out}). 
\begin{equation}
	output =  \textit{Softmax}(o_u + \alpha o_c)
	\label{eq:out}
\end{equation}

The model is trained in an end-to-end manner. More detailed information about the architecture and training can be found in the following section.

\section{Experiments and Evaluation}

\subsection{Dataset}

SARC\footnote{\href{http://nlp.cs.princeton.edu/SARC/}{http://nlp.cs.princeton.edu/SARC/}} \cite{sarc} is a self-annotated corpus for sarcasm detection. SARC is the largest available sarcasm detection dataset for this task and contains more than a million of sarcastic/non-sarcastic samples extracted from Reddit\footnote{\href{https://www.reddit.com/}{https://www.reddit.com/}}. Every instance in SARC is a response to a set of comments. The response is annotated by its author as either sarcastic or non-sarcastic.
In this work, we concatenate all of the available comments for each response in chronological order into a single comment. 

We evaluate our system on the latest version of the balanced SARC (SARC V2.0, Main balanced). Due to the lack of a pre-defined validation set, we randomly hold out $10\%$ of the training set data as our validation set. All hyper-parameters are tuned based on the performance on the validation set. Table~\ref{tab:data:stat} shows the SARC (V2.0) dataset statistics.

\begin{table}[ht]
	\small
	\begin{center}
		\begin{tabular}{l|c|cc}
			\hline
			\multicolumn{2}{l|}{} & non-sarcastic & sarcastic\\ \hline\hline
			\multirow{3}{*}{Train} & Data Size & 128,541 & 128,541 \\
            & \# Avg. Comment & 60.9 & 60.9 \\ 
            & \# Avg. Response & 55.0 & 54.5 \\ \hline
			\multirow{3}{*}{Test} & Data Size & 32,333 & 32,333  \\
            & \# Avg. Comment & 60.8 & 60.8  \\ 
            & \# Avg. Response & 55.8 & 54.7  \\ \hline
			\multicolumn{2}{l|}{Vocabulary} & \multicolumn{2}{c}{95,043}  \\
			\hline
		\end{tabular}
	\end{center}
	\caption{\label{tab:data:stat} SARC main balanced V2.0 statistics.}
\end{table}

The motivation behind using the SARC dataset as our primary benchmark is threefold: (1) SARC is the largest available dataset for sarcasm detection. Consequently, SARC is the most appropriate dataset for training a sophisticated deep-learning based model. Also, due to its size, the evaluation results could be considered more robust and unbiased. (2) SARC is specifically developed to investigate the necessity of contextual information in sarcasm detection in realistic settings. This characteristic aligns well with the motivation of our work. (3) This dataset is author-annotated and has a small false-positive rate for the sarcastic labels \cite{sarc}, thus providing reliable annotations. Importantly, its self-annotation characteristic avoid annotation errors induced by third-party annotators.

\subsection{Experimental Setup}
\label{subsect:setup}	
	We use the pre-trained $300$-$D$ Glove $840B$ vectors \cite{glove} to initialize our word embedding vectors. All hidden states of BiLSTMs for both input encoding and re-reading have $300$ dimensions ($r=300$ and $d=300$). The weights are learned by minimizing the log-loss (Equation~\ref{eq:loss}) on the training data via the Adam optimizer~\cite{adam}. The initial learning rate is 0.0001. To avoid overfitting, we use dropout~\cite{dropout} with the rate of 0.5 for regularization, which is applied to all feedforward connections. During training, the word embeddings are updated to learn effective representations for the sarcasm detection task. We use a fairly small batch size of 32 to provide more exploration power to the model. We consider 200 and 100 as the maximum acceptable length of the comment and response respectively ($n \leq 200$ and $m \leq 100$). In other words, only 200 and 100 words of the given comment and response is processed and the rest (in case of existence) are thrown away.
    
\begin{equation}
	\begin{split}
		y_i^* &= \textit{argmax}(output_i) \\
		l &= - \frac{1}{N} \sum^N_{i=0} (y_i \log (y_i^*) + (1-y_i) \log (1-y_i^*))
	\end{split}
	\label{eq:loss}
\end{equation}

\subsection{Results}

Here we evaluate our model based on two versions of SARC. (1) \citet{cascade} is the most recent work that use SARC dataset for evaluation. It is not clear which version of SARC is used, but they have released their train and test sets\footnote{\href{https://github.com/SenticNet/CASCADE--Contextual-Sarcasm-Detection}{https://github.com/SenticNet/CASCADE--Contextual-Sarcasm-Detection}}. We refer to this dataset as SARC$_{csd}$ in the rest of this paper. In this sub-section (Results), we use SARC$_{csd}$ to compare our system with the reported performances in \citet{cascade}.  (2) We use the SARC V2.0 in next section (Ablation and Configuration Study) to report standard results on SARC V2.0 and compare the performance of different configurations of our model.

\begin{table}[ht]
	\begin{center}
	\small
		\begin{tabular}{lcc}
			\hline
			\multirow{2}{*}{\textbf{Model}}& \multicolumn{2}{c}{\bf{Test Set}} \\ \cline{2-3}
			&\textbf{F1}&\textbf{Acc$^a$}\\ 
			\hline \hline
			(1) Bag of Words  & 64\% & 63\% \\ 
			(2) CNN & 66\% & 65\% \\
            (3) CASCADE $-$ Pf$^b$ & 66\% & 68\% \\ \hline
            (4) CNN-SVM \cite{poria_sarc}  & \textcolor{blue}{68\%} & \textcolor{blue}{68\%} \\
            (5) CUE-CNN \cite{amir_2016}  & \textcolor{blue}{69\%} & \textcolor{blue}{70\%} \\
            (6) CASCADE \cite{cascade} & \textcolor{blue}{77\%} & \textcolor{blue}{77\%} \\ \hline \hline
			(7) \textbf{Ours (AMR)} & \textbf{68\%} &  \textbf{70\%} \\ \hline
            \multicolumn{3}{l}{$^a$\textbf{Acc}: Accuracy} \\ 
            \multicolumn{3}{l}{$^b$\textbf{Pf}: Personality Feature} \\
			\hline
		\end{tabular}
	\end{center}
	\caption{\label{tab:cascade:result} F1-measures and Accuracies of models on the test set of SARC$_{csd}$. The second three (4,5, and 6) models benefit from personality feature (their results are shown in \textcolor{blue}{blue}). Whereas the first three models (1,2, and 3), similar to our model; only rely on response or response and comment. Our models (AMR) achieves the F1-measure and accuracy of $68\%$ and $70\%$ respectively, the best results observed on SARC$_{csd}$ among similar methods which does not use personality features.}
\end{table}

Table~\ref{tab:cascade:result} shows the F1-measures and accuracies of models on the test set of SARC$_{csd}$. The first row shows the results of a baseline classifier using the bag-of-words method. All other listed models are deep learning based. The second model is a simple CNN applied to the given utterance/response. The third system is the CASCADE model \cite{cascade} without using the personality features. This system use the context in a pipeline manner via a discourse feature vector. The next three reported models benefit from stylometric and personality features (The result of such methods are shown in \textcolor{blue}{blue}). 

Bag-of-words approach obtained the lowest performance whereas all deep learning based models outperform it. Among all deep learning ones, the CNN baseline has the lowest performance. The CNN baseline only relies on the given utterance/response highlighting the impact and importance of considering both comment and response in the disambiguation process. 

Comparing methods that benefit from personality features and user profiling (4,5, and 6) with the ones that do not (1,2, and 3), it is clear that such features are very helpful for sarcasm detection. However, user profiling helps a model primarily by providing information about the user's behavior or how the user forms sarcastic sentences. In other words, it does not really enrich the model's capability toward understanding what constructs sarcasm in general. More over, user history and information may not always be available for extracting such features. Importantly, one of the main goals of this work is to move toward solving the sarcasm {\bf understanding} issue in a dialog system. In particular, we are mainly interested in the language understanding aspect of sarcasm detection. As such, we aim to build an end-to-end system that does not depend on any additional information or assumption (user profiling, topic modeling, etc.) other that the sequence of the sentences (the conversation). Due to these considerations, the fair comparison would be comparing the results of our system with the fist three models in Table~\ref{tab:cascade:result}, which demonstrates the effectiveness of our models.  

From Table~\ref{tab:cascade:result} we can see that AMR achieves an F1-measure and accuracy of $68\%$ and $70\%$ respectively on the test set of SARC$_{csd}$, which are the best reported results among the existing comparable baselines for sarcasm detection. Here we obtain $2\%$ improvement on both F1-measure and accuracy on the test data of SARC$_{csd}$ in comparison with the previous state-of-the-art system; CASCADE without personality feature (row 3 in the Table~\ref{tab:cascade:result}). It is interesting to note that although we do not employ user profiling, our performance is similar and competitive with several baselines that use user profiling (CNN-SVM \cite{poria_sarc} and CUE-CNN \cite{amir_2016}).

\begin{table*}[ht]
	\begin{center}
	\small
		\begin{tabular}{lcccc}
			\hline
			\multirow{2}{*}{\textbf{Models}}& \multicolumn{4}{c}{\bf{SARC V2.0 Test Set}} \\ \cline{2-5}
			&\textbf{Precision}&\textbf{Recall}&\textbf{F1-Measure}&\textbf{Accuracy}\\ 
			\hline \hline
			(01) AMR & 69.33\% & 69.64\% & \textbf{69.48\%} & \textbf{69.45\%} \\ \hline
			(02) Conversation-dependent  & 70.23\% & 66.36\% & 68.24\% & 69.11\% \\ 
            (03) Utterance-only  & 70.86\% & 64.66\% & 67.62\% & 69.04\% \\ \hline
            (04) AMR $-$ Attention & 69.39\% & 68.79\% & 69.09\% & 69.22\% \\
			(05) AMR $-$ Re-Reading & 72.93\% & 60.20\% & 65.96\% & 68.93\% \\			
			(06) AMR $-$ Re-Reading $-$ Attention & \textbf{74.76\%} & 55.31\% & 63.58\% & 68.32\% \\ \hline
			(07) AMR $-$ difference & 70.07\% & 67.53\% & 68.78\% & 69.34\% \\
            (08) AMR $-$ element-wise product & 70.41\% & 67.01\% & 68.67\% & 69.42\% \\ 
			(09) AMR $-$ element-wise product $-$ difference & 71.19\% & 65.50\% & 68.23\% & \textbf{69.45\%} \\
            (10) AMR with only element-wise product & 70.75\% & 65.05\% & 67.78\% & 69.07\% \\ \hline
            (11) AMR $-$ train embedding & 67.22\% & \textbf{69.68\%} & 68.43\% & 67.85\% \\ \hline
		\end{tabular}
	\end{center}
	\caption{\label{tab:ablation} Ablation study results. Precision, Recall, F1-Measure, and Accuracy of different models on the test set of SARC V2.0.}
\end{table*}

\subsection{Ablation and Configuration Study}
In this section, we conduct an ablation and configuration study of our model to examine the importance and effect of each major component. We report the performance (Precision, Recall, F1-Measure, and Accuracy) of different variants of our model on the test set of SARC V2.0 in Table~\ref{tab:ablation}.

The first row shows the performance of the proposed model, AMR. Rows 2 and 3 study the impact of the conversation-dependent and utterance-only parts of the models. Rows  4-6 examine the impact of attention and re-reading stages by removing either one (rows 4 and 5) or both components (row 6).  Rows 7-10 investigate the effect of data augmentation in attention and classification of conversation-dependent part of the proposed model. Specifically, we consider removing the different data augmentations shown in Equation \ref{eq:prj:c}, \ref{eq:prj:r}, and \ref{eq:out:conv}.  Finally, row 11 shows the result of our model without fine-tuning the word embedding during the training procedure. 

First, we compare the models based on their F1-Measure and Accuracy. The results show that removing any part of our model leads to reduced test set performance both in terms of F1-Measure and accuracy (expect for row 9 where accuracy remained the same), indicating the usefulness of these components in general. 

We observe that AMR performs noticeably better than both \emph{Utterance-only} and \emph{Conversation-dependent} configurations, validating the intuition of our design. It is noteworthy that \emph{Conversation-dependent} model performs better than the other one, suggesting the importance of considering the conversation and context for this task.
Comparison of rows 4, 5, and 6 suggests that although both of \emph{Attention} and \emph{Re-Reading} are important, but \emph{Re-Reading} has a more significant impact on the performance of AMR. 

A closer look into the precisions and recalls of the different models suggests an interesting trend --- removing different components of the model typically leads to improved precision in sarcasm detection but suffers from significantly reduced recall.  This is evidenced by the results of the first 10 rows.  Comparing the first three rows, it is interesting to note that either part of the model (conversation-dependent or utterance-only) individually achieves slightly higher precision but significantly lower recall. The fact that by combining the two our model was able to achieve significantly improved recall suggests that the two parts were able to detect different types of sarcasms, which is consistent with our intuition.  

Removing fine-tuning of the word embedding during the training has an opposite effect with reduced precision but little or no impact on the recall. This suggests that by fine tuning the word embeddings for the sarcasm detection task, we were able to increase the specificity of the sarcasm detector without sacrificing the sensitivity. 

\section{Analysis}
In this section, we first show visualization of the energy functions (i.e. attention) in the attention stage (Equation \ref{eq:energy}) and its saliency for an instance from the SARC V2.0 test set. Next, we study the performance of our system (Utterance-only, Conversation-dependent and AMR) against the length of comment and response.

\subsection{Attention Study}
\begin{figure}[ht]
	\centering
	\includegraphics[width=.48 \textwidth]{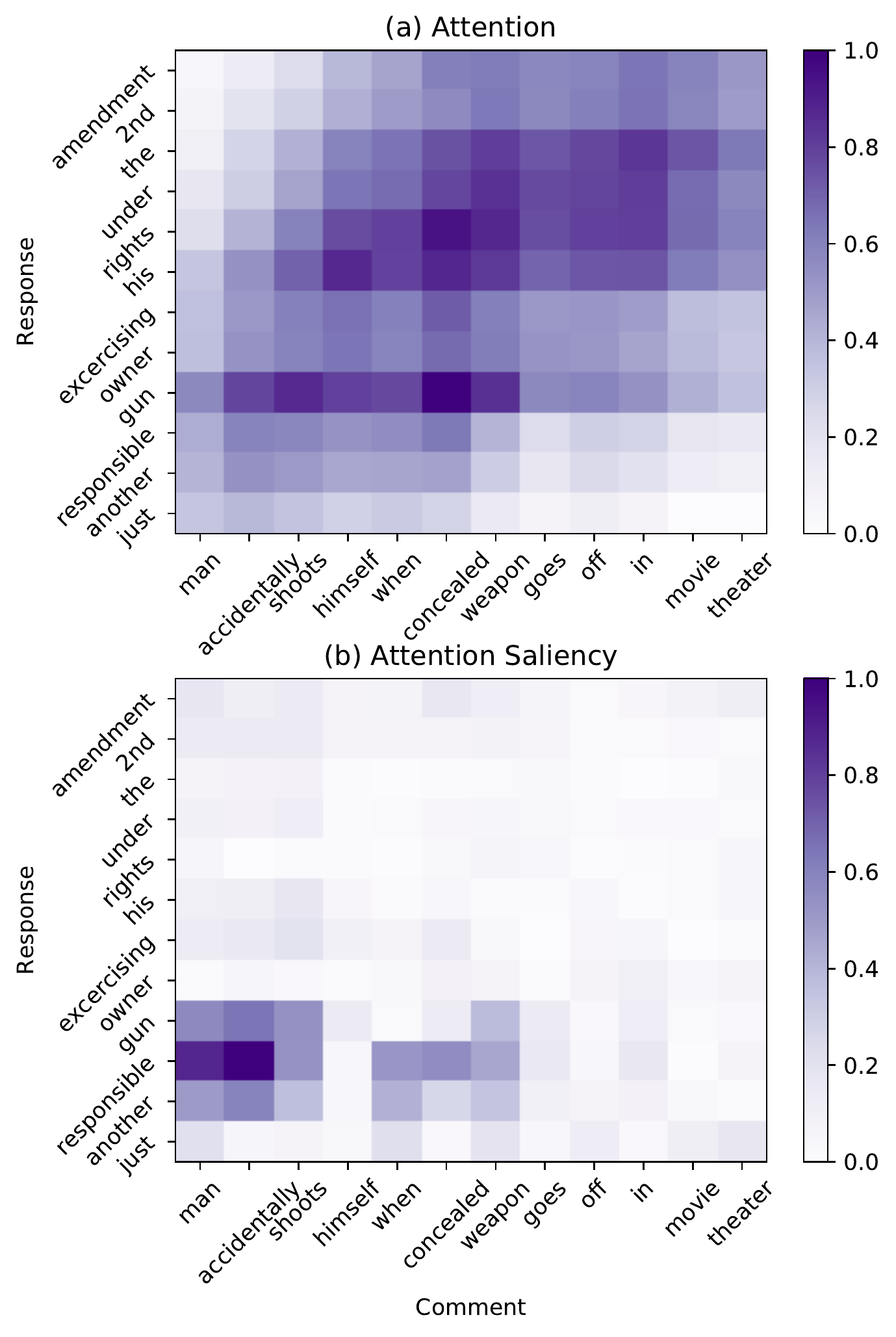}
	\caption{Normalized attention (a, top) and normalized attention saliency (b, bottom) visualization for a sarcastic instance from the test set of SARC V2.0.\label{fig:att}}
\end{figure}

Here we show a visualization of the normalized attention (Equation~\ref{eq:energy}) and normalized attention saliency\footnote{For more details refer to \cite{sali_vis}} in Figure~\ref{fig:att}. 

We show a comment and response pair, where the comment is \textit{``man accidentally shoots himself when concealed weapon goes off in movie theater.''}, and the response is \textit{``just another responsible gun owner exercising his rights under the 2nd amendment.''} which is a sarcastic response and AMR identifies it as sarcastic response as well. Attention visualization in Figure~\ref{fig:att} indicates that the model could successfully attend to relevant pairs of words like $<$gun, shoots$>$, $<$gun, concealed$>$, $<$gun, weapon$>$, $<$his, himself$>$, etc. However, still we cannot clearly explain the model's prediction. Thus we use the attention saliency to visualize the impact of each word pair toward the model's prediction. Attention saliency is the absolute value of the partial derivative of the model prediction respect to the attention. Larger saliency indicates stronger impact on the model's prediction. According to the attention saliency visualization in Figure~\ref{fig:att} (b), the phrase pair of $<$another responsible gun, man accidentally shoots$>$ has the highest impact toward identifying the aforementioned example as sarcastic, which is consistent with human intuition. This demonstrates and verifies the model's ability in understanding comment and response and then utilizing the crucial relationships between the comment and response for identifying sarcastic responses. The word ``responsible'' in the response appears to be the key phrase that deliver the sarcastic intent of the response --- when paired with the phrase ``man accidentally shoots'' we see the highest saliency, suggesting the most significant impact toward the final prediction.

\subsection{Length Study}
One of the advantage of our model is its prediction interpretability. AMR contains two major parts; \emph{Utterance-only} and \emph{Conversation-dependent}. Each part makes its own prediction. Then AMR combines utterance-only and conversation-dependent predictions using a trainable variable $\alpha$ to obtains its final prediction. Consequently, the impact of each part toward the final prediction can be computed. In other words, we can determine which part affects the final prediction the most.

\begin{figure}[ht]
	\centering
	\includegraphics[width=.47\textwidth]{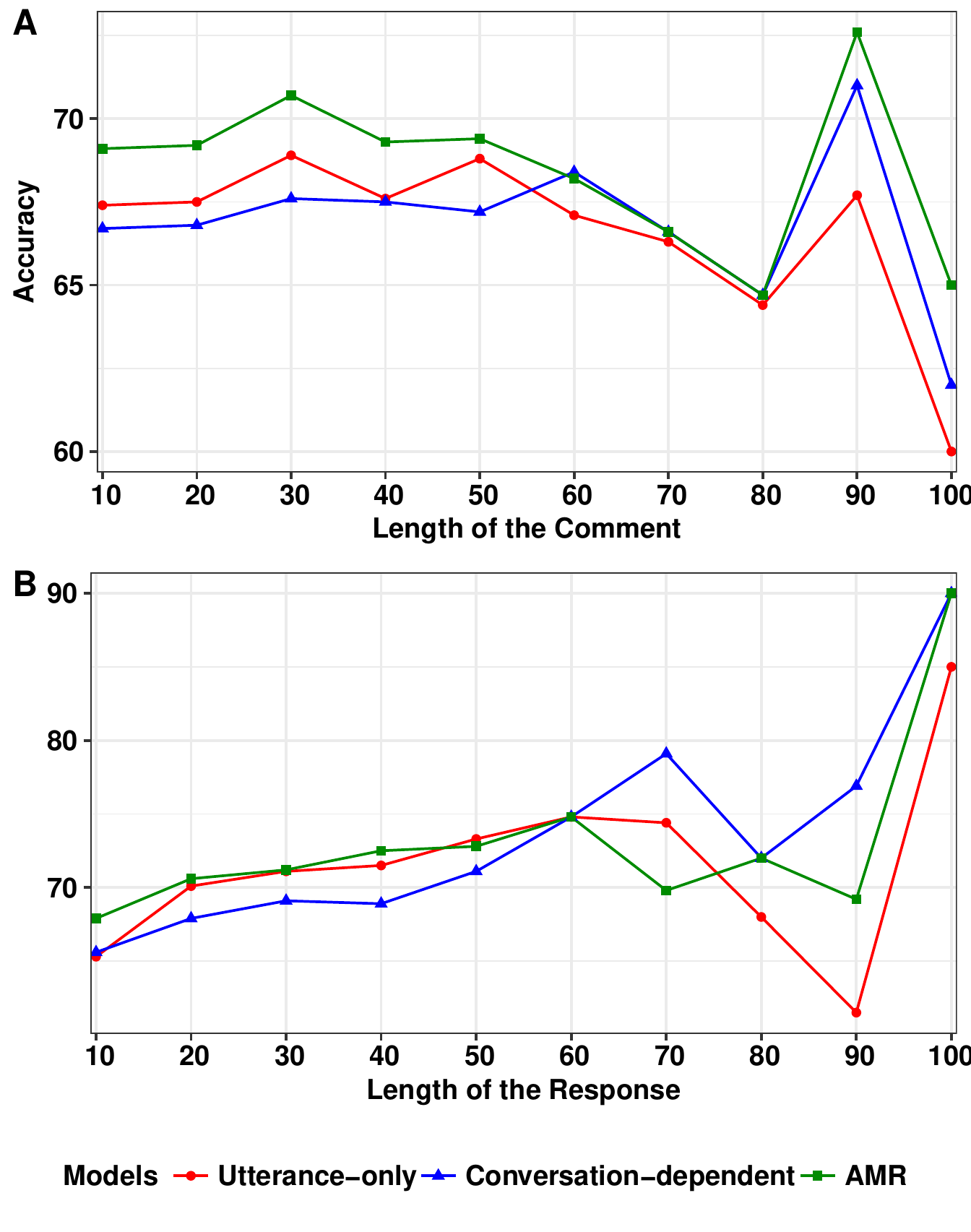}
	\caption{Test accuracy of AMR and its sub-parts (Utterance-only and Conversation-dependent) against the length of the comment (A) and response (B).\label{fig:length:study}}
\end{figure}

Figure~\ref{fig:length:study} depicts the performance of AMR (green line), Utterance-only  part (red line), and Conversation-dependent part (blue lines) against length of the comment (A, left), and length of the response (B, right) respectively.

According to Figure~\ref{fig:length:study}, the utterance-only part provides more accurate predictions for short comments ($n \leq 50$). We believe that the utterance-only part of AMR is capable of automatically extracting useful lexical and grammatical cues from utterance which could be beneficial for detecting sarcastic utterances/responses. Consequently, among samples with short comment; thus less contextual information, the utterance-only part shows better performance. It is noteworthy that the performance of AMR is almost always higher than both utterance-only and conversation-dependent parts. However, the conversation-dependent part performs better for longer comments ($50 < n \leq 200$). This observation is consistent with our expectation because long comments are more likely to have relevant and crucial information for determining the sarcastic intent of the response. Such an analysis verifies the intuition behind the design our model. 

Despite of the plot A in Figure~\ref{fig:length:study}, plot B does not reflect a very coherent behavior and trend among the reported settings. Interestingly, for the very short responses category ($m \leq 10$) which is also the most frequent response category, the conversation-dependent part performs better than the utterance-only part. Due to lack of information in very short responses, disambiguation of such samples are usually reliant on the comment. If we ignore the aforementioned category ($m \leq 10$), plot B illustrates similar behavior and trend for utterance-only and conversation-dependent parts. The utterance-only part perform better for short responses ($10 < m \leq 50$) and the conversation-dependent part beats the utterance-only part for long responses ($50 < m \leq 100$). 

Overall, Figure~\ref{fig:length:study} suggests that the conversation-dependent part performs better when (1) we do not have enough information in the response ($m \leq 10$) or (2) the response or the comment is too long ($n, m > 50$). We believe that in case of dealing with long comment or response, we require some guidance for attending to the important and influential sub-parts of the comment or response. Such a goal can be achieved by utilizing an attention mechanism on both comment and response.

\section{Conclusion}
We propose a novel interpretable end-to-end sarcasm detection model that benefits from both the utterance and the conversational context in parallel. Our evaluations successfully demonstrate the effectiveness of the proposed model. We provide an extensive oblation study that illustrates and justifies the importance and impact of different components of the proposed model. Moreover, we study the model's behavior by visualizing attention and attention saliency. Finally, we present an interesting data analysis to examine the impact of utterance and conversational context on the model's predictions. Our future work will extend our study to include the world fact information in the disambiguation procedure to produce more robust and accurate predictions.

\bibliography{naaclhlt2019}
\bibliographystyle{acl_natbib}
\end{document}